\documentclass[10pt,twocolumn,letterpaper]{article}

\usepackage[pagenumbers]{cvpr} 
\usepackage{graphicx}
\usepackage{amsmath}
\usepackage{amssymb}
\usepackage{booktabs}
\usepackage{tikz}
\usetikzlibrary{tikzmark}
\usepackage{booktabs}
\usepackage{multirow}
\usepackage{xcolor} 
\usepackage{amsmath} 
\usepackage{graphicx} 

\usepackage{amsmath}
\usepackage{algorithm}
\usepackage{algorithmic}
\definecolor{cvprblue}{rgb}{0.21,0.49,0.74}
\usepackage[pagebackref,breaklinks,colorlinks,allcolors=cvprblue]{hyperref}

\def\paperID{4670} 
\def\confName{CVPR}
\def\confYear{2025}

\title{PainterNet: Adaptive Image Inpainting with Actual-Token Attention and Diverse Mask Control}

\author{
Ruichen Wang\\
OPPO AI Center\\
{\tt\small wangruichen@oppo.com}
\and
Junliang Zhang\\
OPPO AI Center\\
{\tt\small 1623204324@qq.com}
\and
Qingsong Xie\\
OPPO AI Center\\
{\tt\small xieqingsong1@oppo.com}
\and
Chen Chen\\
OPPO AI Center\\
{\tt\small chenchen4@oppo.com}
\and
Haonan Lu\\
OPPO AI Center\\
{\tt\small luhaonan@oppo.com}
}

\begin{document}
\maketitle

\begin{abstract}
   Recently, diffusion models have exhibited superior performance in the area of image inpainting. Inpainting methods based on diffusion models can usually generate realistic, high-quality image content for masked areas. However, due to the limitations of diffusion models, existing methods typically encounter problems in terms of semantic consistency between images and text, and the editing habits of users. To address these issues, we present PainterNet, a plugin that can be flexibly embedded into various diffusion models. To generate image content in the masked areas that highly aligns with the user input prompt, we proposed local prompt input, Attention Control Points (ACP), and Actual-Token Attention Loss (ATAL) to enhance the model's focus on local areas. Additionally, we redesigned the MASK generation algorithm in training and testing dataset to simulate the user's habit of applying MASK, and introduced a customized new training dataset, PainterData, and a benchmark dataset, PainterBench. Our extensive experimental analysis exhibits that PainterNet surpasses existing state-of-the-art models in key metrics including image quality and global/local text consistency.
\end{abstract}    
\section{Introduction}
\label{sec:intro}
The rapid progress of diffusion models \cite{ho2020denoising, ho2022classifier} has significantly advanced image generation techniques \cite{chen2023pixart, saharia2022photorealistic, ramesh2022hierarchical}, which are utilized in various applications, including prompt-based conditional editing \cite{brooks2023instructpix2pix, hertz2022prompt}, controllable generation \cite{mou2024t2i, zhang2023adding}, and personalized image synthesis \cite{lu2023specialist, ruiz2023dreambooth, gal2022image}. Among these, image inpainting \cite{xu2023review} is a key application that uses guidance information to restore missing regions in images, allowing users to create content in specified areas based on textual prompts, making it a highly sought-after feature in recent years.

\begin{figure}[t]
\begin{center}
\scalebox{1}{
   \includegraphics[width=1\linewidth]{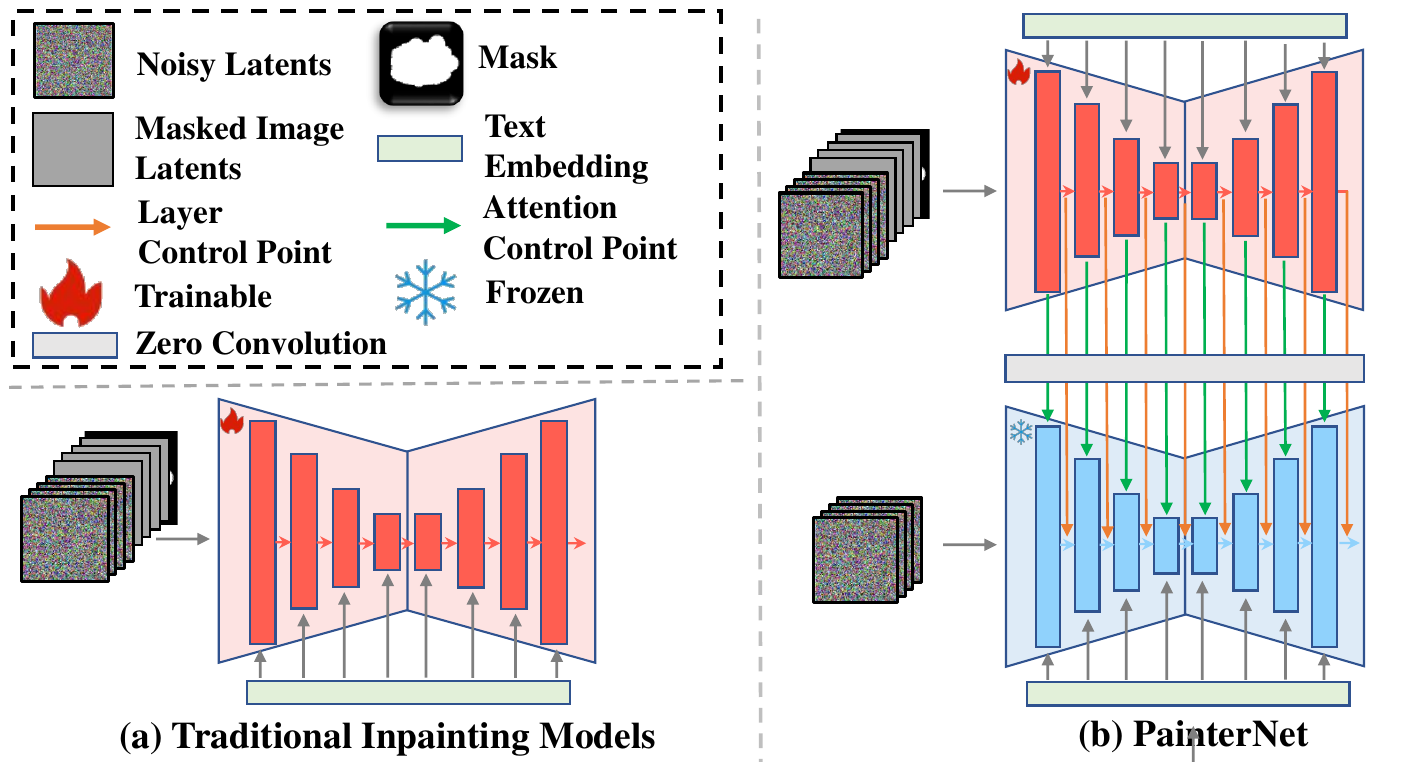}
}
\end{center}
\vspace{-1.0em}
\caption{Comparison of previous inpainting architectures and PainterNet. (a) Dedicated Inpainting model based on diffusion model, enhanced by extended input channels and fine tuning. (b) We propose the plug-and-play approach PainterNet, which introduces an additional branch guidance model for hierarchical dense control via layer and attention control points.}
\label{f_1}
\end{figure}

 Conventional diffusion-based inpainting methods can be divided into two main categories: modifying the sampling strategy \cite{avrahami2023blended, avrahami2022blended, corneanu2024latentpaint, lugmayr2022repaint, zhang2023towards} and utilizing dedicated inpainting models \cite{rombach2022high, wang2023imagen, xie2023smartbrush, yang2023uni}. The former involves sampling from a pretrained diffusion model over masked regions while maintaining the integrity of unmasked areas, which is a training-free approach. However, this method often leads to discontinuities in image generation due to inadequate attention to mask boundaries and contextual information. The latter approach enhances the diffusion model by extending input channels and fine-tuning to effectively address corrupted images and masks (as shown in Fig. \ref{f_1}(a)). While dedicated inpainting models yield improved results, they require fine-tuning of the diffusion backbone and handle both conditioning and generation within a single UNet branch. This not only necessitates extensive data but also limits the model's portability.


\begin{figure*}[t]
\begin{center}\scalebox{1}{
    \includegraphics[width=1\linewidth]{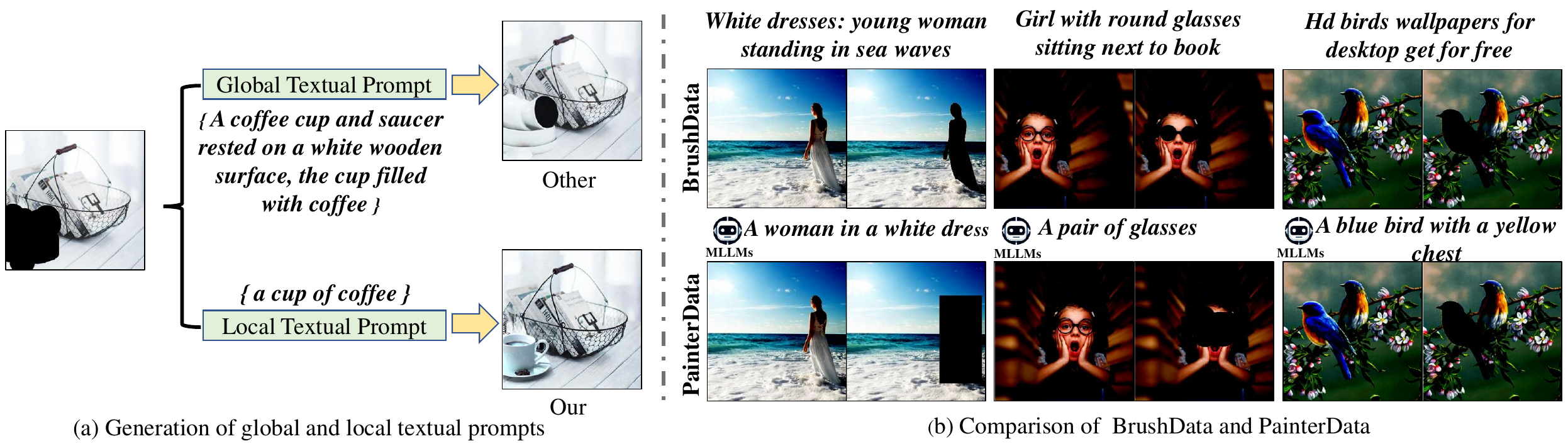}}
\end{center}
\vspace{-1.20em}
\caption{
Comparison of generation and datasets under local and global textual prompts. (a) Our method is able to generate correct results based on the local textual prompt, whereas other methods are not able to ensure consistency of generation when using global textual prompt. (b) In contrast to BrushData, our PainterData contains multiple types of masks (e.g., bounding box, irregular, segmentation-based) as well as local textual prompts generated by a multimodal large language models (MLLMs).
}
\label{f_1.2}
\end{figure*}

Recently, control-based image colorization techniques, such as ControlNet-Inpainting \cite{zhang2023adding} and BrushNet \cite{ju2024brushnet}, have emerged as promising alternatives. These methods incorporate additional control branches to achieve flexible, plug-and-play solutions, and can perform well without the need for heavy training of dedicated inpainting models.
Despite the significant improvements and flexibility offered by existing control-based methods, they still face several challenges. ControlNet-Inpainting \cite{zhang2023adding}  has flaws in terms of image pixel control, making it difficult for inpainted images to remain completely consistent with the original input.
BrushNet \cite{ju2024brushnet} is trained on segmentation mask data(as shown in the first row of Fig. \ref{f_1.2}(b)), which introduces additional information(mask shape) during the training process while  more flexible and personalized mask inputs are expected considering user habit.
 In addition, all the above methods usually rely on global textual prompts that do not provide localized detail descriptions, which may lead to inconsistencies between the generated local content and the expected prompts (as shown in Fig. \ref{f_1.2}(a)).

To address the limitations of existing methods, we propose a novel image editing framework—PainterNet.
Specifically, inspired by BrushNet \cite{ju2024brushnet}, we introduced an additional branch that employs a layered integration approach, progressively incorporating full UNet features into a pre-trained UNet, enabling dense pixel-wise control (as shown in Fig. \ref{f_1}(b)). 
However, in order to obtain more detailed descriptions of the mask area from the input prompt, we adopted local textual prompts as the input. Further more, we introduced Attention Control Points (ACP) and Actual-Token Attention Loss (ATAL) to make the model focus more on the mask area based on the information provided by the input local textual prompts.

We have also constructed a new training dataset, PainterData, and a corresponding benchmark, PainterBench, based on BrushData \cite{ju2024brushnet}. Considering the actual applications of users, We designed a diversified mask generation strategy to fit the types of masks that may appear in the actual use of the inpainting model (including boundary box and irregular scribbles to simulate fingers,
as shown in the second row of Fig. \ref{f_1.2}(b)).
In addition, we leveraged multimodal large language models (MLLMs), such as ShareGPT \cite{chen2023sharegpt4v}, to generate local textual prompts for the masked regions, encouraging the model to focus on generating localized content.
The contributions of our work are summarized as follows:
\begin{itemize}
\item[$\bullet$] We proposed a novel plug-and-play image inpainting framework, PainterNet, which introduces an additional branch for layered and dense pixel-wise control, enhancing the generation capabilities of the diffusion model.
\end{itemize} 
\begin{itemize}
\item[$\bullet$] We introduced Attention Control Points (ACP) and the Actual-Token Attention Loss (ATAL) to capture the semantic associations between masked images and local textual prompts. This ensures that our model can utilize the information in the local textual prompt well to generate the missing area in the image.
\end{itemize} 
\begin{itemize}
\item[$\bullet$]We constructed a new dataset pipeline, PainterData, which automatically generates local textual prompts using multimodal large language models, and design diverse mask generation strategy. This enables the model to better understand the semantic information of local regions, making it more applicable to real-world scenarios.
\end{itemize} 

\begin{figure*}
\begin{center}\scalebox{0.9}{
    \includegraphics[width=1\linewidth]{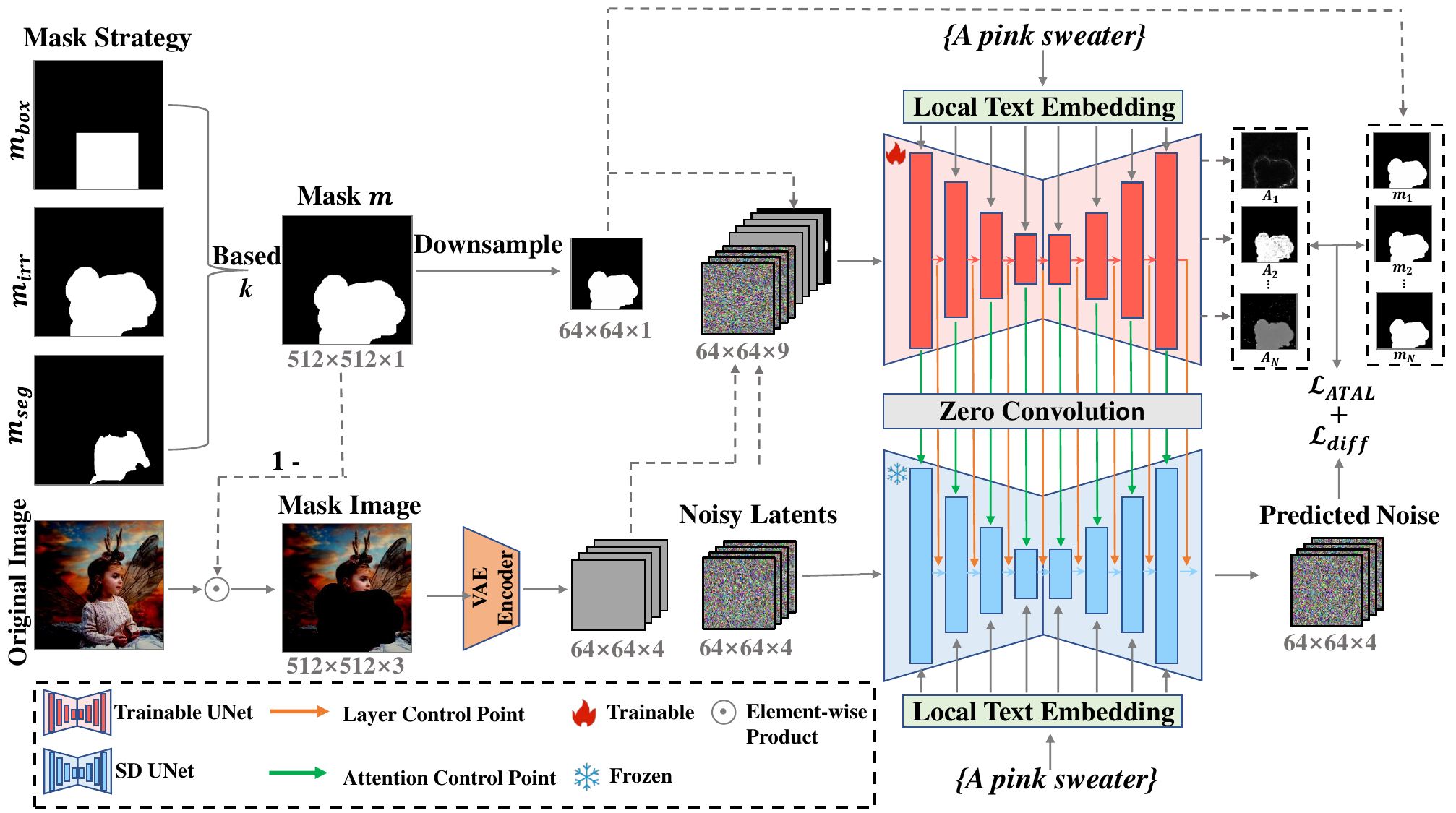}}
\end{center}
\vspace{-1.20em}
\caption{Overview of our method.
Our PainterNet introduces an additional branch that uses a hierarchical approach to gradually incorporate the complete UNet features into the pre-trained UNet layer by layer through layers and attentional control points. Meanwhile, we designed the Actual-Token Attention Loss (ATAL) $\mathcal{L}_{\text {ATAL}}$ to direct the model's attention to the mask region. Masking Strategy generates diverse masks (e.g., bounding box $m_{box}$, irregular $m_{irr}$, and segmentation-based $m_{seg}$) and selects the input mask shape based on a random number $k \in [0, 1]$. $A_{i}, i \in [1,2,...N]$ represents the cross-attention map of the $i$-th layer, where $N$ is the total number of layers. $m_{i}$ denotes the mask $m$ resized to fit $A_{i}$. $\mathcal{L}_{\text {diff}}$ denotes diffusion loss.
}
\label{f_2}
\end{figure*}
\section{Related Work}
\label{sec:formatting}
\subsection{Diffusion Models}

In recent years, diffusion-based image generation techniques \cite{avrahami2022blended, ho2020denoising, li2024blip} have gradually gained significant attention as a research field. This surge of interest is primarily due to the superior image generation quality of diffusion models compared to traditional generative models (such as GANs \cite{goodfellow2020generative, karras2017progressive, karras2019style}). 
Therefore, with the introduction of diffusion-based models, the quality of text-to-image generation has significantly improved. DALL-E 2 \cite{ramesh2022hierarchical} utilizes CLIP \cite{radford2021learning} to achieve text-to-image mapping through a diffusion mechanism and trains a CLIP decoder. Imagen \cite{saharia2022photorealistic}, on the other hand, leverages large pretrained language models like T5 \cite{raffel2020exploring} to achieve exceptional alignment between images and text using textual data. Stable Diffusion \cite{rombach2022high} employs efficient encoding in latent space to generate images with rich details and diverse styles. 
Notably, stable diffusion is one of the most popular open-source text-to-image generation methods, and several different versions have been developed, ranging from stable diffusion v1.5 (SD 1.5) to stable diffusion v2.0 (SD 2.0), and then to stable diffusion XL (SD XL) [28]. Each version has demonstrated significant improvements in image fidelity and generation speed.
Additionally, in downstream applications like generating anime styles \cite{counterfeit-v3-huggingface}, Van Gogh styles \cite{huggingface_vangogh}, and specific roles \cite{civitai_ironman}, diffusion models demonstrate remarkable style adaptability and high-quality generation capabilities.

Our PainterNet is also built on the foundation of stable diffusion models, implementing hierarchical dense control to fully leverage their capability for generating high-fidelity images. It can also be easily adapted to various downstream tasks, providing plug-and-play flexibility.





\subsection{Image Inpainting}
Given a masked scene image, the objective of image inpainting is to recover the occluded regions in a natural and plausible manner \cite{quan2024deep,xu2023review}. In the early stages of development, most deep learning methods were based on paradigms such as autoencoders \cite{peng2021generating, zheng2019pluralistic}, autoregressive transformers \cite{wan2021high}, and  GAN-based paradigms  \cite{sargsyan2023mi, xu2023image, zhao2021large,zheng2022image}. These methods typically relied on auxiliary handcrafted features, resulting in suboptimal performance. Recently, techniques based on diffusion models \cite{lugmayr2022repaint} have gained widespread attention for their exceptional ability to generate high-quality images \cite{ho2020denoising,rombach2022high}.

Image inpainting also benefits from text-guided techniques based on diffusion models. For a given pretrained diffusion model \cite{avrahami2022blended, avrahami2023blended}, a sampling strategy that replaces the latent unmasked regions with the noisy versions of the known areas during the sampling process can produce satisfactory images for simple inpainting tasks. However, these methods fail to adequately perceive information at the mask boundaries and the context of the unmasked regions, leading to results that lack coherence. 
Previous methods \cite{meng2023distillation, wang2023imagen,xie2023smartbrush,xie2023dreaminpainter,yang2023uni} addressed this issue by fine-tuning the proposed content-aware and shape-aware models. Specifically, 
SmartBrush \cite{xie2023smartbrush} combines text and shape guidance to enhance the diffusion U-Net with object mask prediction, leading to better perception of mask boundary information. 
Stable Diffusion Inpainting \cite{rombach2022high} fine-tunes the diffusion model specifically for the inpainting task, where the input to the U-Net consists of a mask, a masked image, and noisy latent variables. 
HD-Painter \cite{manukyan2023hd} is built upon Stable Diffusion Inpainting, enhancing generation quality in painting tasks through attention layers and attention guidance. 
However, these methods face challenges in effectively transferring their inpainting capabilities to arbitrary pretrained models, which limits their applicability.

Recently, to enable any diffusion model with inpainting capabilities, the community has fine-tuned ControlNet \cite{zhang2023adding} for image inpainting. Ju et al \cite{ju2024brushnet}. proposed BrushNet, characterized by its plug-and-play and content-aware properties. However, ControlNet has limitations in understanding the perception of masks and masked images. BrushNet, trained with global prompts, lacks local detail descriptions, and its branch removes the cross-attention layers, resulting in a lack of semantic understanding in hierarchical control, which may lead to inconsistencies between the generated images and the text prompts. 
Our model, PainterNet, innovatively proposed a solution for local textual prompt input and has been trained with superior model design and data.
\section{Method}
Our proposed PainterNet, shown in Fig.\ref{f_2}, is designed to precisely capture detailed information in masked regions during image generation. To this end, we adopt a dual-branch strategy to embed mask information, leveraging hierarchical dense control through layer control points and Attention Control Points (ACP). Additionally, we introduce a Actual-Token Attention Loss (ATAL) that directs the model's focus to the masked regions, ensuring alignment between text prompts and generated image content. 
In addition, to meet the high-fidelity demands of realistic mask shapes and local semantics, we propose a novel data construction pipeline. This pipeline combines local textual prompts with diverse masks (e.g., bounding box, irregular, segmentation-based), effectively training our model for improved performance.


\subsection{Preliminaries}
In this paper, we employ Stable Diffusion(SD) \cite{rombach2022high} as the foundational model for image restoration. The model takes a text prompt $P$ as input and generates the corresponding image $x_{0}$. Stable Diffusion comprises three main components: an autoencoder $(\mathcal{E}(\cdot), \mathcal{D}(\cdot))$, a CLIP text encoder $\tau(\cdot)$, and a U-Net $\epsilon_{\theta}(\cdot)$. Typically, the model is trained under the following diffusion loss constraint:
\begin{equation}\label{eq_1}
\mathcal{L}_{\text {diff }}=\mathbb{E}_{z_{0},  \epsilon \sim \mathcal{N}(0,1), t, \boldsymbol{c}}\left[\left\|\epsilon-\epsilon_{\theta}\left(z_{t}, t, \boldsymbol{c}\right)\right\|_{2}^{2}\right],
\end{equation}
where $\epsilon \sim \mathcal{N}(0,1)$ is the randomly sampled Gaussian noise, $t \in[1, T]$ is the time step, $T$ is the total time step, $\boldsymbol{c}=\tau(P)$ is text embedding, $z_{0}=\mathcal{E}\left(x_{0}\right)$ is the latent representation of $x_{0}$, and $z_{t}$ is calculated by $z_{t}=\alpha_{t} z_{0}+\sigma_{t} \epsilon$ with
the coefficients $\alpha_{t}$ and $\sigma_{t}$ provided by the noise scheduler.

\subsection{PainterNet}
To help the model better capture masked image information and achieve high-quality image inpainting, we designed PatinerNet based on a diffusion model. PatinerNet introduces an additional branch copied from the pretrained Stable Diffusion U-Net, as shown in Fig.\ref{f_2}. 
Not like BrtushNet \cite{ju2024brushnet} that excludes its cross-attention layers, we keep all the U-Net structure but only change the input dims from $4$(for $z_{t}$) to $9$($4$ for $z_{t}$, $4$ for masked image latent $z_{0}^{\text {m }}$ and $1$ for downsampled mask $m$). The $z_{0}^{\text {m}}$ is extracted using the same VAE model from Stable Diffusion. Moreover, instead of using global textual prompts, we utilize local textual prompts for cross-attention layers both in SD U-Net and PatinerNet-Branch. So we have $\boldsymbol{c}_{l}=\tau(P_{l})$, where $P_{l}$ is the local textual prompt.Unlike most existing control-based methods that only use layer control points, we further proposed Attention Control Points(ACP) to apply directly influence on cross-attention layers as the global textual prompts has been changed to local ones.

Similar to ControlNet \cite{zhang2023adding}, we employ zero convolution layers to connect the frozen model with the trainable PatinerNet, avoiding noise interference during the early stages of training. The features of PatinerNet are gradually inserted into the frozen diffusion model, enabling pixel-level fine-grained control:





\begin{gather}\label{eq_2}
attn_{i} = \epsilon_{\theta,attn}^{\text {PN }}\left(\left[z_{t}, z_{0}^{\text {m}}, m\right], t,\boldsymbol{c}_{l}\right)_{i} \notag \\
lay_{i} = \epsilon_{\theta,lay}^{\text {PN }}\left(\left[z_{t}, z_{0}^{\text {m }}, m\right], t,\boldsymbol{c}_{l}\right)_{i} \notag \\
{\epsilon^{\prime}}_{\theta,attn}\left(z_{t}, t, \boldsymbol{c}_{l}\right)_{i}=\epsilon_{\theta,attn}\left(z_{t}, t, \boldsymbol{c}_{l}\right)_{i}+ w \cdot \mathcal{Z}_{attn}\left(attn_{i}\right)  \notag \\
{\epsilon^{\prime}}_{\theta,lay}\left(z_{t}, t, \boldsymbol{c}_{l}\right)_{i}=\epsilon_{\theta,lay}\left(z_{t}, t, \boldsymbol{c}_{l}\right)_{i}+ w \cdot \mathcal{Z}_{lay}\left(lay_{i}\right) \notag \\
\end{gather}
where $\epsilon_{\theta,attn}^{\text {PN }}$, $\epsilon_{\theta,lay}^{\text {PN }}$ denote the cross-attention output and layer output in layer $i (i \in[1, N])$ of PainterNet $\epsilon_{\theta}^{\text {PN }}$, where $N$ is the number of layers. And $\epsilon_{\theta,attn}$, $\epsilon_{\theta,lay}$ denote the cross-attention input and layer input in layer $i$ of SD U-Net $\epsilon_{\theta}$.
$\text { [} \cdot \text {] }$ refers to the concatenation operation, and $\mathcal{Z}_{attn}, \mathcal{Z}_{lay}$ are zero-convolution operations. $w$ is the preservation scale used to adjust the influence of PainterNet on the pre-trained diffusion model.

\subsection{Actual-Token Attention Loss}

During the training of diffusion models, the diffusion loss $\mathcal{L}_{\text {diff }}$ ensures the model's ability to generate content, but it lacks explicit alignment constraints between the textual prompts and the masked region pixels, which can lead to issues of prompt neglect. Therefore, we propose ACtual-Token Attention Loss (ATAL) to seamlessly decompose the cross-attention features in PainterNet and enforce the attention to focus on the masked regions without adding extra modules.

Specifically, given a text embedding $\boldsymbol{c}_{l}$ and a latent $z_{t}$, PainterNet projects $z_{t}$ and $\boldsymbol{c}_{l}$ to form queries $Q_{i}$ and keys $K_{i}$ computing the cross-attention map and flattening the textual information into spatial features:
\begin{equation}\label{eq_3}
A_{i}=\operatorname{Softmax}\left(\frac{Q_{i} K_{i}^{T}}{\sqrt{d_{i}}}\right),
\end{equation}
where $A_{i} \in \mathbb{R}^{HW \times L} $ is the cross-attention map of layer $i$ of PainterNet, $H$ and $W$ denote the height and width of the image, and $L$ denotes the length of the text encoding ($L=77$ in CLIP \cite{radford2021learning}). 
We first define the indices of the actual input text tokens as $S$, which are the index numbers less than the actual token length from the text embedding, and then removing the starting and ending special tokens (i.e., \textless SOT \textgreater and \textless EOT \textgreater). Then, we use actual textual prompts in the response region of the attention map to direct the model's attention to the corresponding masked region:
\begin{equation}\label{eq_4}
\mathcal{L}_{\mathrm{ATAL}}=\frac{1}{N} \sum_{i=1}^{N}\left\| \frac{1}{L_{S}} \sum_{j\in S}A_{i,j}-m_{i}\right\|_{2}^{2},
\end{equation}
where $L_{S}$ denotes the length of $S$. $A_{ij} \in \mathbb{R}^{HW \times 1 }$ represents the $j$-th actual text token in the cross-attention map of the $i$-th layer in PainterNet.   $m_{i} \in \mathbb{R}^{HW \times 1 }$ denotes the mask resized from $m$ to fit the size (HW) of $A_{ij}$.
Finally, we achieve high-quality image inpainting by combining the diffusion model loss with Actual-Token Attention Loss, promoting consistency between local generation and textual prompts. Our overall loss function is as follows:
\begin{equation}\label{eq_5}
\mathcal{L}=\mathcal{L}_{\text {diff}}+\beta \mathcal{L}_{\text {ATAL}}
\end{equation}
where $\beta$ is a hyperparameter.

\subsection{PainterData}

To the best of our knowledge, most publicly available datasets for training image editing models consist of global text prompts along with corresponding random brush masks or segmentation-based masks, such as BrushData \cite{ju2024brushnet}. This setup requires models to perform local edits based on global text prompts, lacking detailed descriptions of the regions to be edited, which can result in outputs deviating from the expected results. Additionally, user-generated masks are often characterized by randomness and personalization, with their shapes and sizes significantly differing from predefined segmentation masks. Therefore, models need to be capable of handling more flexible and non-standard masks to better accommodate diverse user requirements.

To address these challenges, we propose a new dataset pipeline called PainterData. Specifically, we modify the BrushData \cite{ju2024brushnet} dataset by replacing global captions with localized captions generated by pre-trained large-scale language models (e.g., ShareGPT \cite{chen2023sharegpt4v}) and post-processing the generated captions with ChatGLM \cite{team2024chatglm}.
This involves extracting the main objects from the captions and creating shorter, object-specific captions, thereby generating localized captions for the dataset.

In addition, to accommodate the randomness of user-provided masks, we introduce a new training mask generation strategy that allows users to provide either fine (e.g., segmentation-based masks $m_{seg}$) or coarse (e.g., bounding-box $m_{box}$ or irregular scribbles to simulate fingers $m_{irr}$) masks. 
Specifically, we generate rectangular or square masks based on the segmentation masks from BrushData \cite{ju2024brushnet}, or simulate finger-like irregular scribbles by applying dilation followed by random scribbling. During training, we select mask shapes based on a random number $k \in [0, 1]$, reducing the model's sensitivity to segmentation mask shapes and enhancing its versatility. 
Formally, given a segmented base mask $m_{seg}$, we can obtain the final mask:

\begin{equation}\label{eq_2}
\begin{array}{rl}
m = 
\begin{cases} 
m_{box} & \text{if } k \leq 0.25 \\[8pt]
m_{irr} & \text{if } 0.25 < k \leq 0.75\\[8pt]
m_{seg} & \text{other}
\end{cases}
\end{array}
\end{equation}
In this way, the model adapts to different mask shapes, allowing the user to input mask shapes more easily without having to strictly follow the contours of the target instance. Thus, our proposed PainterData compensates for the lack of localized detail description in BrushData \cite{ju2024brushnet} and is more suitable for practical applications. Specific mask generation strategies can be found in the supplementary material.

\section{Experiments}
\subsection{Experimental Setup}
\noindent \textbf{Benchmark.}
Currently, commonly used datasets in the field of image synthesis include CelebA \cite{liu2015deep}, CelebA-HQ \cite{huang2018introvae}, ImageNet \cite{deng2009imagenet}, MSCOCO \cite{lin2014microsoft}, and Open Images \cite{kuznetsova2020open}. However, these datasets are not suitable for training and evaluating image synthesis methods based on diffusion models due to issues such as small focused regions or low quality. The recently proposed BrushBench \cite{ju2024brushnet} benchmark is specifically designed for image synthesis methods based on diffusion models. 
However, the captions in BrushBench  are global text prompts, which overlook local detail descriptions. Moreover, most of the masks in BrushBench are segmentation-based, which can be challenging for users to accurately draw object masks during usage. Therefore, BrushBench also overlooks the practical applications of inpainting in real-world scenarios.

To address this gap, we introduce PainterBench for image inpainting with various mask shapes encountered in real-world scenarios. Specifically, we generate different mask shapes based on the image data from BrushBench using our generative strategy. Additionally, the textual prompts in PainterBench are generated as localized prompts through a pre-trained large-scale language model, such as ShareGPT \cite{chen2023sharegpt4v}. Furthermore, the dataset ensures a uniform distribution across different categories, including humans, animals, indoor scenes, and outdoor scenes. This balanced allocation facilitates fair evaluation across categories, promoting better evaluation fairness. Detailed PainterBench can be found in the supplementary material.


\noindent \textbf{Metrics.}
For quantitative analysis, we utilize Image Reward (IR) \cite{xu2024imagereward}, a human preference evaluation model for text-to-image tasks; Aesthetic Score (AS) \cite{schuhmann2022laion}, a linear model based on real image quality evaluations; CLIP Similarity (CLIP Sim) \cite{wu2021godiva}, which measures text-image consistency between the globally generated image and the corresponding text prompt; Local CLIP Similarity (Local CLIP Sim) \cite{wu2021godiva}, which assesses text-image consistency between the generated image in the masked region and the corresponding text prompt; and Gdino Accuracy (Gdino Acc) evaluates the accuracy of the model in local generation. Specifically, we extract the generated regions from the image based on the mask, then use the grounding dino (Gdino) \cite{liu2023grounding} model to obtain the predicted boxes with local text prompts as input.
We calculate whether each predicted phrase is consistent with the input local text prompt, thereby obtaining the accuracy of local generation.

\begin{figure*}[!t]
\begin{center}\scalebox{0.78}{
    \includegraphics[width=1\linewidth]{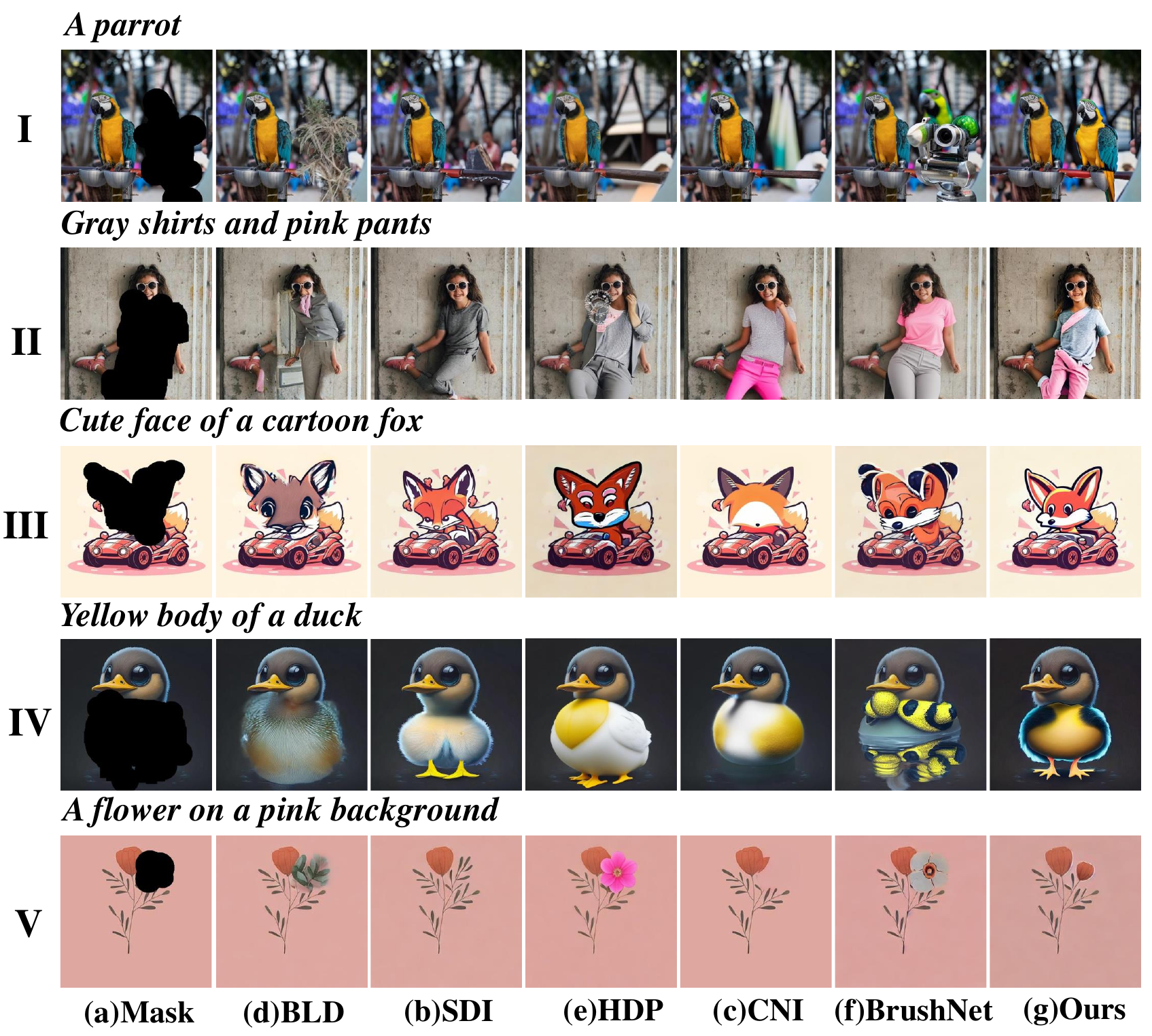}}
\end{center}
\vspace{-1.20em}
    \caption{Comparison of the performance of PainterNet and previous image drawing methods in various styles of drawing tasks: I, II for nature images, III and IV for cartoons, and V for illustrations. }
\label{f_3}
\end{figure*}

\begin{figure*}[t]
\begin{center}\scalebox{0.74}{
    \includegraphics[width=1\linewidth]{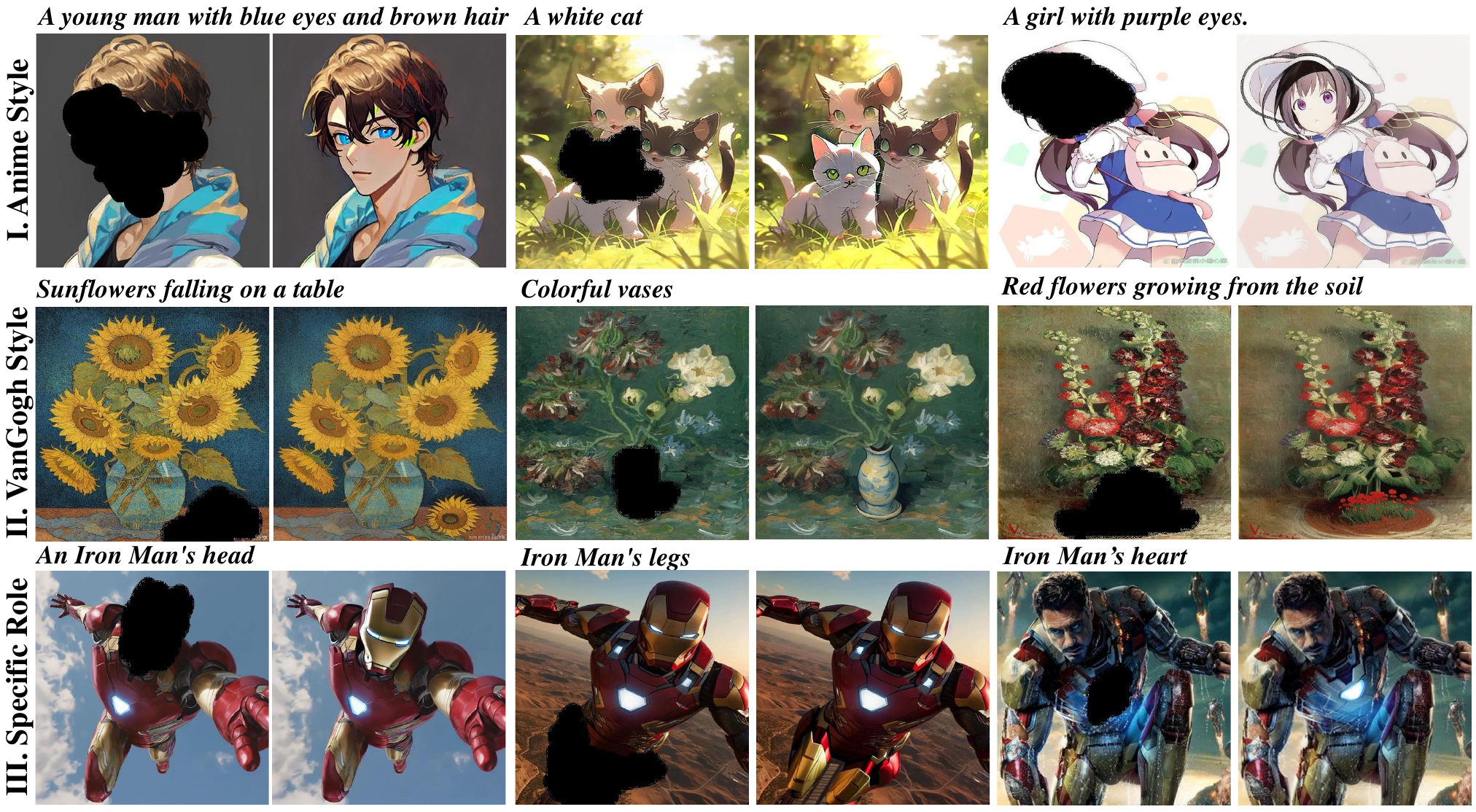}}
\end{center}
\vspace{-1.20em}
\caption{Generative effects of our PainterNet migration to other downstream models. Model I generates anime style outputs \cite{counterfeit-v3-huggingface}, Model II produces VanGogh style art \cite{huggingface_vangogh}, and Model III generates specific roles (such as Iron Man) \cite{civitai_ironman}.}
\label{f_4}
\end{figure*}

\begin{table*}[!ht]
\centering
    \caption{Quantitative comparisons among PainterNet and other diffusionbased inpainting models in PainterBench. SD indicates that Stable Diffusion V1.5 was used as the base model. SD XL indicates that Stable Diffusion XL was used as the base model. \textbf{Bold} denotes the best. \underline{Underline} denotes the second best. }

    \setlength{\tabcolsep}{2.5mm}{
    \begin{tabular}{cc|ccccc}
    \toprule[1pt]
    \specialrule{0em}{1pt}{1pt}
    \multicolumn{2}{c|}{Method}                         &IR          & AS       & CLIP Sim & Local CLIP
Sim  & Gdino Acc  \\ 
    \specialrule{0em}{1pt}{1pt}
    \hline
    \multirow{6}{*}{\rotatebox[origin=c]{90}{SD}}

      &BLD \cite{avrahami2023blended}        & 0.89   & 6.40      &25.64  &22.27  &0.93     \\
      &SDI \cite{rombach2022high}        & 1.00   & 6.34      &25.78  &19.01  &0.82     \\
    & HDP \cite{manukyan2023hd}        & \underline{1.09}        & \underline{6.49}      &\underline{25.92}  &22.13  &0.94     \\
    &CNI \cite{zhang2023adding}        & 0.51   & 6.32      &24.82  &21.66  &\underline{0.94}     \\
     &BrushNet \cite{ju2024brushnet}        & 0.87        & 6.34      &25.47  &\underline{22.35}  &0.93    \\
     &Ours                               &\textbf{1.12}         & \textbf{6.51}     &\textbf{26.12}  &\textbf{22.67}  &\textbf{0.96}     \\    

    \hline
    \multirow{3}{*}{\rotatebox[origin=c]{90}{SD XL}}
     &SDXLI \cite{podell2023sdxl}                       & \textbf{1.25}         & \textbf{6.34}     &\underline{26.71}  &19.49  &0.84   \\
     &BrushNet \cite{ju2024brushnet}      & 0.88        & 5.91    &26.22  &\underline{22.37}  &\underline{0.94}   \\

    &Ours                               & \underline{1.17}         & \underline{6.15}     &\textbf{26.78}  &\textbf{23.06}  &\textbf{0.95}    \\ 

    \bottomrule[1pt]
    \end{tabular}
    }
    \label{t_1}
\end{table*}

\begin{table*}[!ht]
\centering
    \caption{Ablation of different compositions on PainterNet.  ACP: Attention Control Point. ATAL: Actual-Token Attention Loss. \textbf{Bold} denotes the best.}

    \setlength{\tabcolsep}{3.5mm}{
    \begin{tabular}{c|ccccc}
    \toprule[1pt]
    \specialrule{0em}{1pt}{1pt}
    Module                         &IR          & AS       & CLIP Sim & Local CLIP
Sim  & Gdino Acc  \\ 
    \specialrule{0em}{1pt}{1pt}
    \hline
    ControlNet-base        & 1.06   & 6.42      &25.81  &22.40  &0.94   \\

    $+$ PainterNet-Branch ($w/o$ ACP)      & 1.08        & 6.40      &25.82  &22.53  &0.94     \\

    $+$ PainterNet-Branch       & 1.12      & 6.42   &25.98 &22.50  &0.95     \\
    
    $+$ ATAL                                 & \textbf{1.12}         & \textbf{6.51}     &\textbf{26.12}  &\textbf{22.67}  &\textbf{0.96}     \\

    \bottomrule[1pt]
    \end{tabular}
    }
    \label{t_2}
\end{table*}

\subsection{Implementation Details}
Unless otherwise specified, we perform inference for different image inpainting methods under the same settings: using NVIDIA L40S GPUs, following their open-source code, and employing Stable Diffusion v1.5 and Stable Diffusion XL as the base model with 50 steps and a guidance scale of 7.5. For fair experimental comparison, we use the recommended hyperparameters for inference according to each method. In our approach, PainterNet and all ablation models are trained for 500 thousands steps on 4 NVIDIA L40S GPUs. For comparisons on PainterBench, we use PainterNet trained on PainterData, with the parameter $\beta$ set to 0.00001. It is important to note that our PainterData utilized only 8\% of the data from BrushData, amounting to 500,000 data.
\subsection{Quantitative Comparison}
Tab. \ref{t_1} provides a quantitative comparison on PainterBench, where we evaluate the inpainting results of various methods, showcasing the performance differences across approaches. As shown in Tab. \ref{t_1}, PainterNet excels in all key metrics, particularly in Gdino Acc (0.96) and Local CLIP Sim (22.67), surpassing all other models. This highlights PainterNet's superior ability to maintain consistency between image content and text prompts, along with its efficiency in restoring fine details, especially in masked regions.

To evaluate performance at higher resolutions, we compared SDXL-based models. As shown in the lower half of Tab. \ref{t_1}, PainterNet continues to perform strongly across multiple metrics. While it scores slightly lower in IR and AS compared to SDXL-inpainting \cite{podell2023sdxl}, this may be due to our method prioritizing alignment between masked areas and text prompts over overall visual quality. Nonetheless, PainterNet still outperforms state-of-the-art methods on three other key metrics, particularly achieving a Local CLIP Sim of 23.06. It’s worth noting that when SDXL is used as the base model, all methods show improvements in Local CLIP Sim, indicating that SDXL offers better text-to-image translation capabilities, which PainterNet leverages effectively for inpainting.

\subsection{ Qualitative Comparison}

Fig.\ref{f_3} presents a qualitative comparison between PainterNet and other leading methods across various scenarios. It demonstrates that PainterNet holds significant advantages in detail restoration, text consistency, and boundary transitions. In Example I, the inpainting result for "a parrot" shows that PainterNet can generate a natural and detailed parrot that seamlessly blends with the unmasked regions. In contrast, most other methods struggle to properly restore the area, often generating background elements in the masked region (e.g., SDI \cite{rombach2022high} , HDP \cite{manukyan2023hd}, and CNI \cite{zhang2023adding}). Additionally, Example III, featuring a "cartoon fox," further validates PainterNet’s exceptional performance in maintaining text consistency. PainterNet’s generated details align more closely with the prompt, whereas other methods tend to produce blurred or incorrect details, problems that PainterNet successfully avoids. This success is due to our dual-branch design, which allows PainterNet to better perceive background information and maintain coherence with the surrounding context.


\subsection{Ablation Study}

We conducted ablation studies to evaluate the impact of each module in PainterNet on overall performance, as shown in Tab. \ref{t_2}. Adding the control branch to the base model resulted in a significant performance boost, particularly in Local CLIP Sim (from 22.40 to 22.53). This indicates that the control branch enhances the model's ability to perceive local features, improving detail restoration in masked areas. After introducing the Attention Control Point (ACP), the model focused more on global image consistency, which led to improvements in overall CLIP Sim (from 25.82 to 25.98) and other metrics, such as IR and AS. However, as the model shifted focus to aligning global features, attention to local regions slightly decreased, causing a minor drop in Local CLIP Sim. Finally, incorporating Actual-Token Attention Loss (ATAL) enabled PainterNet to achieve optimal performance across all key metrics. This demonstrates the effectiveness of ATAL in guiding the model to focus on highly responsive features in masked areas, significantly enhancing both the detail and consistency of local inpainting.

\subsection{ Flexible Migration Capability}
We explored the plug-and-play capabilities of PainterNet to assess its transferability and generalization performance across different downstream models. We tested three stylistically distinct models: I. Anime Style \cite{counterfeit-v3-huggingface}, II. Van Gogh Style \cite{huggingface_vangogh}, and III. Specific Roles (e.g., Iron Man) \cite{civitai_ironman}, with results shown in Fig. \ref{f_4}.

In the anime style, PainterNet generated character images with clear outlines and vivid colors, successfully retaining the artistic essence of anime while accurately reflecting the text descriptions. For the Van Gogh style, PainterNet effectively captured the distinctive brushstrokes and color schemes characteristic of Van Gogh, especially when dealing with complex backgrounds, resulting in artworks with a high degree of artistic coherence and detail. 
Similarly, in generating specific roles, the generated images accurately reproduce the stylistic details and iconic features (e.g., head, legs, etc.) of a specific character (e.g., Iron Man) while matching the user's textual prompts.
These results demonstrate PainterNet's exceptional flexibility in stylistic control and robust generalization capabilities across various styles, highlighting its broad practical application potential.

\section{Conclusion}
This paper proposes a novel image restoration framework, PainterNet, providing more intuitive and excellent inpainting performances. By introducing a dual-branch structure and freezing the original SD Unet branch during training, PainterNet ensures plug-in capability, making it flexible for various DM models. Moreover, the proposed Attention Control Point (ACP) and Actual-Token Attention Loss (ATAL) further enhance the model's focus on masked areas, significantly improving the quality and consistency of generated images. Finally, to support the training and evaluation of PainterNet, we have constructed a new dataset - PainterData, characterized by diverse mask generation strategies and localized prompts, as well as the PainterBench benchmark used to evaluate model performance. Through comprehensive experimental verification, we demonstrated that PainterNet performs excellently in multiple restoration tasks, surpassing existing methods in terms of semantic alignment and detail preservation.

{
    \small
    \bibliographystyle{ieeenat_fullname}
    \bibliography{main}
}


\clearpage
\setcounter{page}{1}
\maketitlesupplementary

\section{Construction of the dataset}
\subsection{Generation of the local textual prompts}
Due to the use of global text prompts in BrushData \cite{ju2024brushnet}, detailed descriptions of masked regions cannot be provided, which may lead to inconsistencies between local generation and text prompts. To address this issue, we employ a multimodal large language model along with corresponding post-processing operations to generate local text prompts for the masks, replacing the global text prompts and ensuring consistency between the generation of masked regions and the text prompts.

\begin{figure*}[!ht]
\begin{center}\scalebox{0.89}{
    \includegraphics[width=1\linewidth]{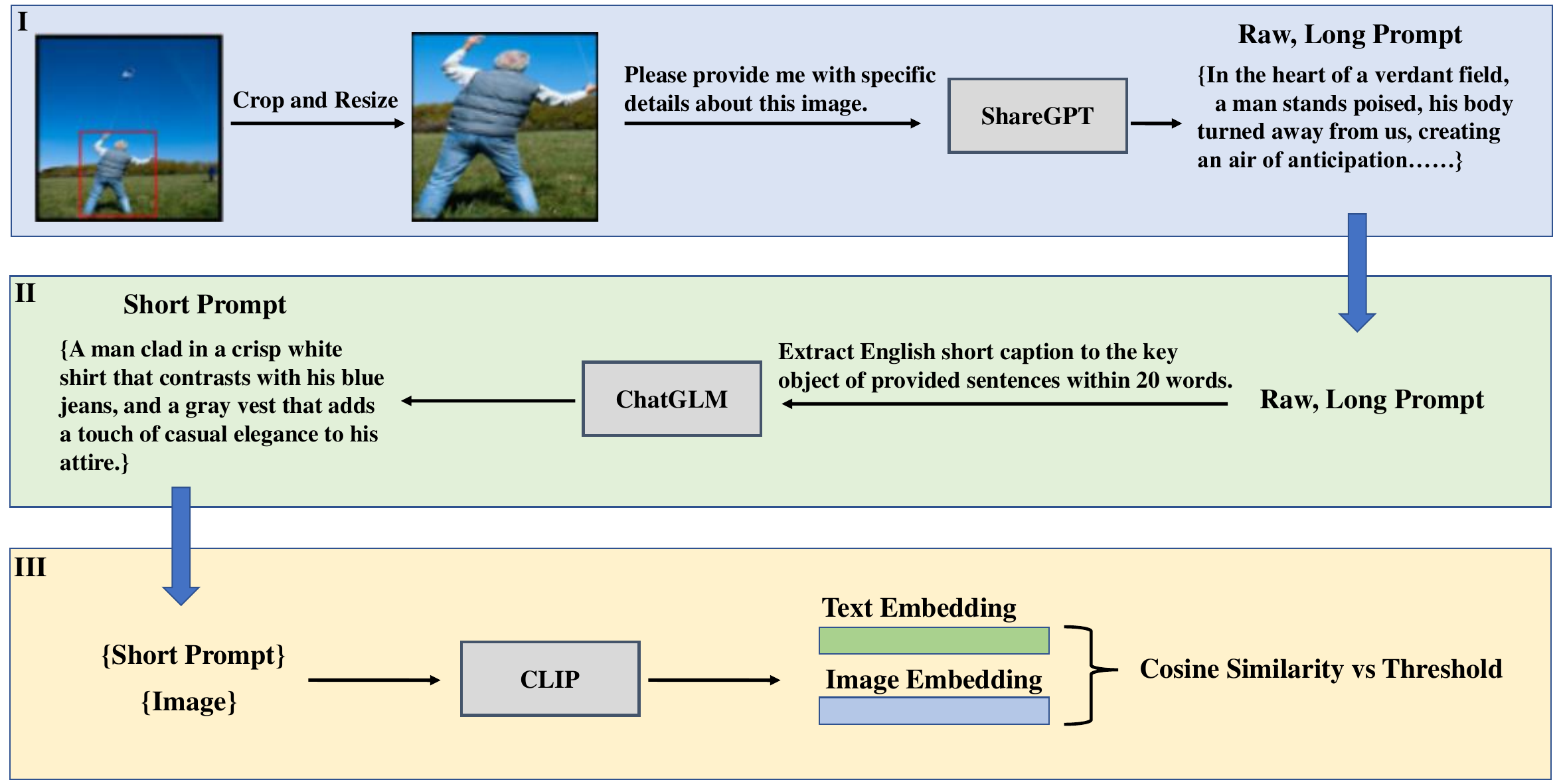}}
\end{center}
\vspace{-1.20em}
\caption{Generation of the local textual prompts. We first obtain the object location, crop it and input it into a multimodal large language model (e.g., ShareGPT \cite{chen2023sharegpt4v}) to obtain the local textual prompts. Then post-processing is performed by ChatGLM \cite{team2024chatglm} and CLIP \cite{radford2021learning}.}
\label{f_7.2}
\end{figure*}

As illustrated in Fig. \ref{f_7.2}, we first obtain the object locations using the segmentation masks provided by BrushData, cropping them and inputting them into a multimodal large language model (e.g., ShareGPT \cite{chen2023sharegpt4v}) to obtain local text prompts (as shown in I of Fig. \ref{f_7.2}). However, some generated prompts may be overly verbose; thus, we utilize ChatGLM \cite{team2024chatglm} to extract concise descriptions of the main objects, resulting in shorter local text prompts (as shown in II of Fig. \ref{f_7.2}). 
Finally, we use CLIP \cite{radford2021learning} to obtain the cosine similarity between the image and the shorter local prompt, and retain the prompt if the similarity exceeds a threshold of 0.2 (as shown in III of Fig. \ref{f_7.2}).
\subsection{Generation of masks}
\label{sec:rationale}
During our training process, we dynamically generate masks using a random number $k \in [0,1]$. These masks include the segmentation mask $m_{seg}$, the bounding box mask $m_{box}$, and the irregular masks $m_{irr}$ to simulate fingers. The goal is to enhance the model's generalization capability and robustness, allowing it to better adapt to the requirements of real-world applications.

\begin{figure*}[!ht]
\begin{center}\scalebox{0.89}{
    \includegraphics[width=1\linewidth]{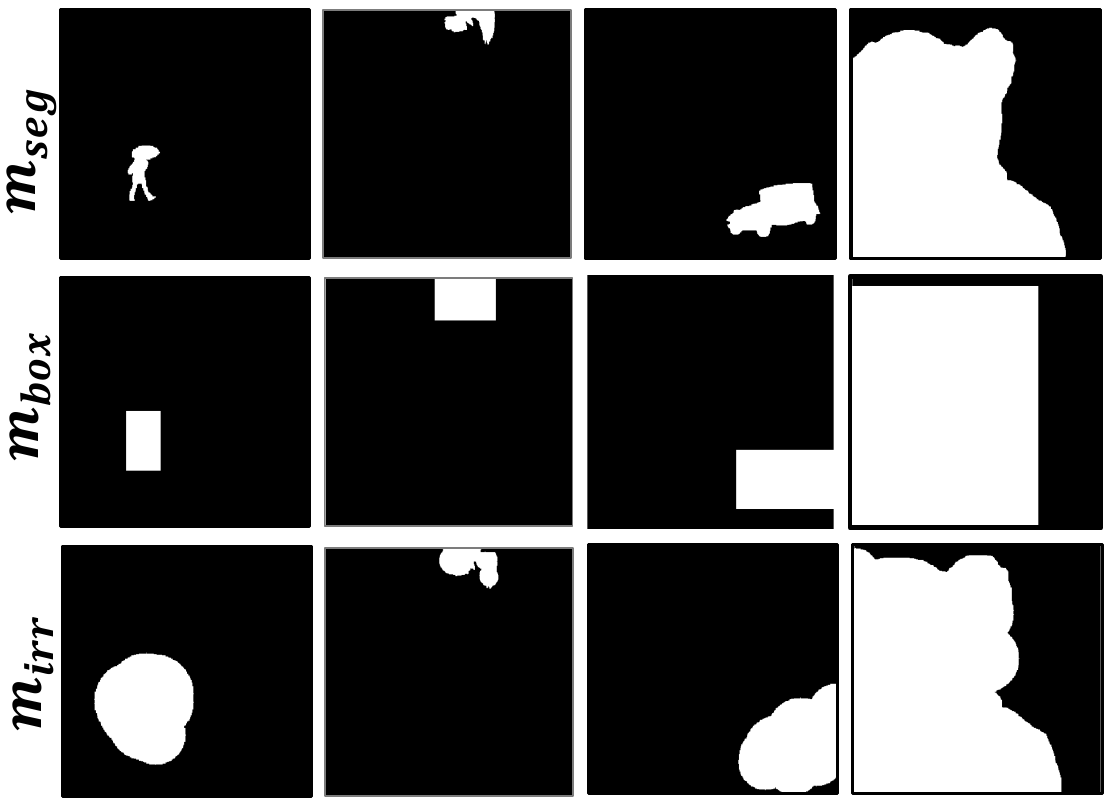}}
\end{center}
\vspace{-1.20em}
\caption{Different mask shapes. Our mask generation strategy can generate diverse masks, which include segmentation-based $m_{seg}$, bounding box $m_{box}$, and irregular $m_{irr}$.}
\label{f_5.1}
\end{figure*}

For the generation of $m_{seg}$, we utilize segmentation masks provided by BrushData, as illustrated in the first row of Fig. \ref{f_5.1}. To create $m_{box}$, we calculate a bounding box around the non-zero pixel locations in the segmentation mask and randomly expand the size of this box. This adjustment increases the diversity of the bounding mask, as shown in the first row of Fig. \ref{f_5.1}. For $m_{irr}$, we apply a dilation operation using a convolutional kernel of random size on the mask. This dilation enhances the continuity of the mask and reduces the separation of targets in the image. Subsequently, we randomly draw lines, circles, or squares on the mask. The quantity and shapes of the drawn elements are controlled by input parameters, further increasing the diversity and complexity of the masks, as depicted in the third row of Fig. \ref{f_5.1}. Details of the generation of $m_{irr}$ are described in Algorithm \ref{A1}.
\section{PainterBench}
\begin{figure*}[!h]
\begin{center}\scalebox{1}{
    \includegraphics[width=1\linewidth]{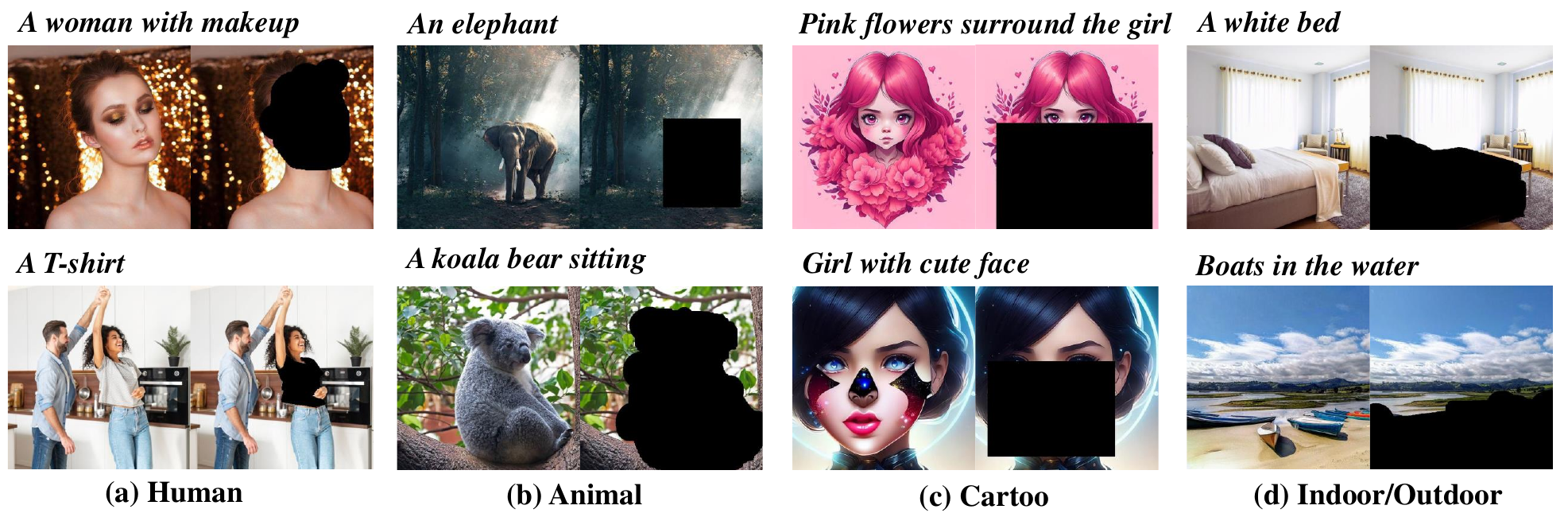}}
\end{center}
\vspace{-1.20em}
\caption{PainterBench overview. Our PainterBench ensures a uniform distribution of categories such as humans, animals, cartoons, and indoor and outdoor scenes.}
\label{f_5.2}
\end{figure*}

As illustrated in Fig. \ref{f_5.2}, our PainterBench ensures a uniform distribution among various categories, including humans, animals, cartoons, as well as indoor and outdoor scenes. This balanced allocation facilitates fair evaluations across different categories, enhancing assessment fairness. Furthermore, our masks also encompass a variety of sizes and shapes, making PainterBench more representative of real-world evaluation scenarios.


\begin{algorithm}[!ht]
\caption{Processes for the generation of $m_{irr}$}
\begin{algorithmic}[1]
\STATE \textbf{Input:} Segmentation-base mask $m_{seg}$
\STATE Compute image dimensions: $h, w \gets \text{shape of } m_{seg}$
\STATE Calculate image total pixels: $t_{img} \gets h \times w$
\STATE Compute $m_{seg}$ total pixels: $t_{m} \gets sum(m_{seg}$)\\
\STATE Compute coverage ratios: $r \gets t_{m}/t_{img} $\\
\textcolor{gray}{\#Applying Dilatation}\\
\STATE Calculate dilation parameters based on $r$
\STATE Generate dilatation kernel $k$ and iterations $iter$ based on $r$
\STATE Generate dilation mask $m_{d}$: \\
$m_{d} \gets cv2.dilate(m_{seg}, k, iter)$\\
\textcolor{gray}{\#Applying Drawing}\\
\STATE Find non-zero points in $m_{d}$
\IF{no points found}
    \STATE $m_{irr} = m_{d}$

    \RETURN $m_{irr}$
\ENDIF
\STATE Determine drawing times based on random selection
\FOR{each drawing iteration}
    \STATE Select a random starting point from found points
    \FOR{each sub-iteration}
        \STATE Generate random angle, length, and brush width
        \STATE Draw line/circle/square on $m_{d}$
        \STATE $m_{irr} = m_{d}$
    \ENDFOR
\ENDFOR
\STATE \textbf{Output:} $m_{irr}$
\end{algorithmic}
\label{A1}
\end{algorithm}



\end{document}